# An explainable machine learning-based approach for analyzing customers' online data to identify the importance of product attributes

Authors: Aigin Karimzadeh, Amir Zakery[1], Mohammadreza Mohammadi, Ali Yavari

## Abstract

Online customer data provides valuable information for product design and marketing research, as it can reveal the preferences of customers. However, analyzing these data using artificial intelligence (AI) for data-driven design is a challenging task due to potential concealed patterns. Moreover, in these research areas, most studies are only limited to finding customers' needs. In this study, we propose a game theory machine learning (ML) method that extracts comprehensive design implications for product development. The method first uses a genetic algorithm to select, rank, and combine product features that can maximize customer satisfaction based on online ratings. Then, we use SHAP (SHapley Additive exPlanations), a game theory method that assigns a value to each feature based on its contribution to the prediction, to provide a guideline for assessing the importance of each feature for the total satisfaction. We apply our method to a real-world dataset of laptops from Kaggle, and derive design implications based on the results. Our approach tackles a major challenge in the field of multi-criteria decision making and can help product designers and marketers, to understand customer preferences better with less data and effort. The proposed method outperforms benchmark methods in terms of relevant performance metrics.

## 1- Introduction

As communication technology develops, the Internet has become a daily tool for many users to explore and exchange information. One of the ways that users interact with the Internet is through online platforms, such as Amazon, and app stores, that collect customer data for various products and services (Abkenar et al., 2021; Angelopoulos et al., 2021). Consequently, this kind of data has become essential in diverse research areas (Heidemann et al., 2012; Li et al., 2023; Nilashi et al., 2023; Noori, 2021). In product design research, some studies have leveraged user data to explore new possibilities and customer knowledge for product development (Bashir et al., 2017; Jeong et al., 2019; Li et al., 2021; Purnama & Masruroh, 2023; Roelen-Blasberg et al., 2023; Wang et al., 2023) . In marketing research, this kind of data can help us to optimize the performance of products and monitor the satisfaction of customers (Bhimani et al., 2019; de Camargo Fiorini et al., 2018). For instance, customers evaluate products online with numeric ratings and text reviews, which are displayed as summary statistics on E-commerce web pages. These ratings reveal valuable information about customer satisfaction, requirement product

---

features, and choices. These feedbacks can influence the decision-making processes of potential customers, market researchers, and product developers (Çalı & Baykasoğlu, 2022). Moreover, determining the preferred product attributes and their benefits, is essential for designing and developing competitive products as they have a vital role in understanding consumer preferences and satisfaction (Kalro & Joshipura, 2023). Therefore, analyzing online customer data is a crucial issue for developing a new product concept, as it reflects the changing preferences and behaviors of online shoppers (Guo et al., 2017).

Furthermore, data-driven product design is a promising approach for developing future products, which are more connected and responsive to user needs (Briard et al., 2023). However, despite being acknowledged as an important approach of design for the upcoming years (Kim et al., 2017), data-driven design is limited and far from being mature yet in product development. Therefore, there is a research gap and a need for data-driven methodologies and suitable guidelines to facilitate data-driven product development (Briard et al., 2023). To address this gap, we present a game theory machine learning (ML) method that streamline product development process using online data. Our method combines three main concepts ML, genetic algorithm and SHAP (SHapley Additive exPlanations) model.

ML is a branch of artificial intelligence (AI) that learns from data and improves its performance without explicit programming. ML can analyze complex and large-scale online data that are beyond human or traditional methods, and automate the data analysis process with self-learning algorithms. Genetic algorithm is a method for finding approximate or near-optimal solutions for optimization problems that can be used as a wrapper method for feature selection. For the game theory part, we use SHAP model, which explains the output of any ML model with game-theoretic Shapley values and assigns scores to each input feature based on its contribution. To clarify, a genetic algorithm is used as a wrapper method with a regressor as the evaluation criterion. We used several other regressors to find out which one is more accurate. These regressors were machine learning algorithms such as K-Nearest Neighbors (KNN), Decision Tree (DT), and so on. The genetic algorithm selects a subset of features in each iteration and evaluates it with one of these algorithms. The genetic algorithm tries to find the optimal subset of features that maximizes the performance of these algorithms according to an evaluation criterion, such as Mean Squared Error (MSE), Mean Absolute Error (MAE), R-squared, or Root Mean Squared Error (RMSE). After the best subsets are selected in a loop, the SHAP model is used to determine the value of each feature in the best subset of features. We use the overall product ratings and the attribute characteristics as inputs for our model.

By using SHAP model, we could provide a suitable guideline for assessing the importance of each attribute of products for the total satisfaction. Our method also derives design implications by analyzing the features and specifications of the products based on the SHAP values. Furthermore, this study addresses a key problem in multi-criteria decision making (MCDM) literature: finding the impact of product attributes on customer satisfaction and the importance of criteria. Existing methods have limitations in dealing with uncertainty and require the performance score of each attribute(Çalı & Baykasoğlu, 2022). All in all, the results can help product designers and marketers to understand customer preferences better with less data and effort.

The main contributions of this article to the literature are as follows: (1) using a genetic algorithm for finding preferred subsets of product features; (2) building an explainable machine learning model that estimates customer ratings and explains the model output with SHAP values; and (3) analyzing SHAP results for comprehensive product design implications.

The article is organized as follows: Section 2 reviews related works. Section 3 introduces methods and concepts. Section 4 presents the implementation and the results and Section 5 compares them with other methods. Section 6 concludes and gives implications.

# 2- Literature review

This study proposes a data-driven product design method based on online customer satisfaction data using game theory and machine learning (ML). In this section, we first review the related literature and identify their limitations, and then we introduce the technical challenges and the ML techniques to address them.

## 2-1 Evaluation of product development

The rapid and unpredictable changes in the market and technology require new methods and techniques for developing and launching successful new products (Briard et al., 2023). Data collection and usage are essential in this dynamic environment, but traditional methods, such as questionnaires or interviews, have limitations, such as subjectivity, low reliability, high cost, and long time (Bi, Liu, Fan, & Zhang, 2019; Çalı & Baykasoğlu, 2022; Groves, 2006; Yang et al., 2019; Yang et al., 2023). In contrast, Online customer data provides a convenient and rapid way to measure customer satisfaction, as customers voluntarily and freely generate their own content and data, unlike surveys (Joung et al., 2021). However, Online data is rich, complex, diverse, and dynamic, which poses difficulties for traditional methods to process and understand. Machine Learning (ML), a subset of Artificial Intelligence (AI), is concerned with computer programmes that can access data and reduce technical issues related to data quality, availability, integration, and scalability and of course Human mistakes. With AI strategies firms can improve their products and services and prepare for the future. AI can analyze massive amounts of market data and predict what the preferences of a customer are (Haleem et al., 2022). Due to the above mentioned argument, it is crucial and very effective to examine and analyze online customer-generated data using machine learning methods for product development comprehensively. ML can also learn from data and improve over time, while traditional methods may depend on human intervention and feedback. Moreover, companies face challenges in integrating data into their product development stages due to the lack of tools and resources (Briard et al., 2023). Therefore, this study proposes a new machine learning methodology to enhance data-driven product design and development, based on the arguments and needs discussed above.

## 2-2 Product development based online data

Customer satisfaction leads to higher purchase intention (Dash et al., 2021). To improve customer satisfaction and loyalty, it is essential to study the effects of product attributes on customer satisfaction (Wang et al., 2018). Customer satisfaction can be measured through online data (Hou et al., 2019). These data can provide design implications for new products by various methods, such as identifying features of customer interests (Park & Kim, 2024). By finding out which product features are important, the manufacturer can learn what most consumers want and need, and use this knowledge to innovate and improve its product (Zhou et al., 2023). Several studies have used different methods to analyze the impact of product attributes on overall customer satisfaction, ranging from conventional techniques like surveys to relatively new techniques like online reviews. For example, Kim and Noh (2019) found the major factors influencing washing machine design using text analysis from online reviews, and performed linear regression to estimate their effect on customer satisfaction. Du et al. (2022) proposed a novel method to assess user satisfaction of smart, connected products (SCPs) based on online reviews. The method uses text mining, sentiment analysis, and decision-making techniques to extract product attributes, evaluate their importance, and analyze user satisfaction degrees. Imtiaz and Ben Islam (2020) analyzed the product reviews on mobile phones to find out how product attributes affect customer satisfaction. They compared five machine-learning algorithms and three feature selection algorithms to discover an optimal set of 21 smart phone attributes that are crucial for customer satisfaction. Suryadi and Kim (2019) used a method that combines word embedding and clustering to find product features, and then uses sentiment analysis and regression to link them to sales rank. Joung and Kim (2023) analyzed customer reviews to extract the product attributes that customers mentioned in each review. In the next step, the importance of product features on customer satisfaction was determined using sentiment analysis and the SHAP model. They then grouped the reviews based on the similarity of customer preferences and importance values for each attribute. Çalı and Baykasoğlu (2022) identified the importance of product features on customer satisfaction (ratings) using a Bayesian based approach. However, most of these studies have deficiencies in identifying and ranking the most important product features that affect customer satisfaction by considering the interactions among them. In other words, there is a research gap and a need for automated AI methodologies and suitable guidelines to determine the significance and characteristics of various features, as well as the optimal combination of features that can facilitate data-driven product development. To address this gap, we present a game theory ML method that uses genetic algorithms to select, rank, and combine product features that can maximize customer satisfaction. We use the overall product ratings and the attribute characteristics as inputs for our model.

## 2-3 The correlation between customer satisfaction and product features

The effect of product features on customer satisfaction is not always straightforward, as previous research has shown (Bi, Liu, Fan, & Cambria, 2019; Deng et al., 2010). The Kano model is a commonly used framework, which categorizes product features into five types such as attractive, performance, must-be, and reverse based on how they influence customer satisfaction (Kano, 1984). For instance, some features are attractive, meaning that they increase customer satisfaction when they are present, but do not decrease it when they are absent. Other features are must-be, meaning that they decrease customer satisfaction when they are absent, but do not increase it when they are present.

Machine learning can detect these nonlinear patterns, but it cannot explain the contribution of product features to customer satisfaction (Joung & Kim, 2023). To overcome this challenge, we use SHAP (SHapley Additive exPlanations), which is a method that can reveal how product features affect customer satisfaction. Furthermore, this study addresses a key problem in multi-criteria decision making (MCDM) literature: finding the impact of product attributes on customer satisfaction and the importance of criteria. Existing methods have limitations in dealing with uncertainty and there is a gap in finding the weights when only the overall satisfaction scores of alternatives are known (Çalı & Baykasoğlu, 2022). We use SHAP to provide a suitable guideline for assessing the importance of each attribute for the total satisfaction. Our method also derives design implications by analyzing the features and specifications of the products based on the SHAP values.

# 3-Method

This section proposes a methodology for using online product ratings to examine how product attributes influence customer satisfaction. The method procedure, illustrated in Fig. 1, has three main phases. First, the online data is preprocessed to make it suitable for analysis. Second, a genetic algorithm identifies the optimal subsets of product attributes that have an impact on user ratings. It employs a machine learning algorithm that predicts user ratings based on subsets of product attributes and evaluates their correlation with the actual ratings as a fitness function. Third, a SHAP model determines the importance of each product attribute in the optimal subsets on the user rating, based on the fitness function from the previous phase.

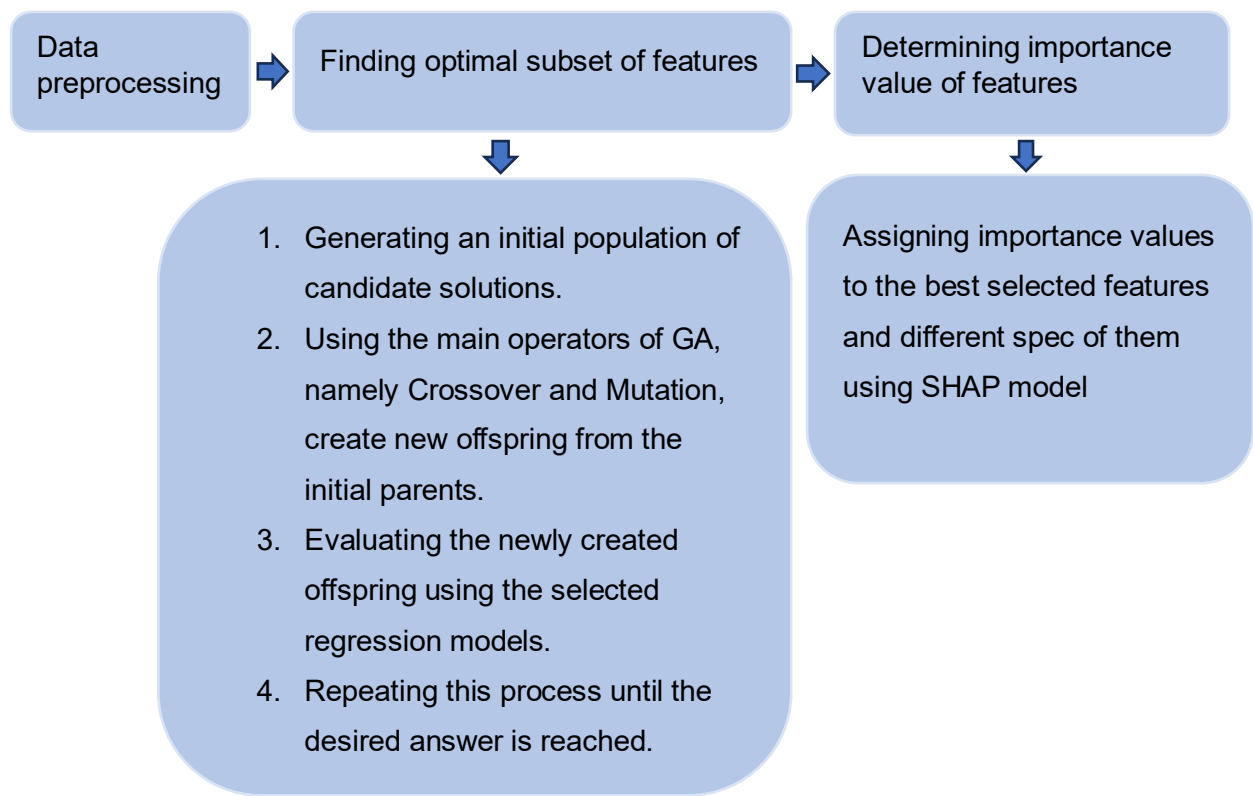

Figure 1: Method procedure

## 3-1-Data preprocessing

In the developed approach, to examine the impact of product features on customer satisfaction, we used a dataset of online product ratings from an e-commerce websites. The dataset contains various kinds of a specific type of product and their features, as well as the user ratings for each product. This ensures that the customers are representative and that we have a large number of ratings to analyze.

For preprocessing, we normalized the numeric features, encoded the categorical features, and handled the missing values. We also created a new feature that combined the number of ratings and the star ratings scores, dropped some irrelevant or redundant features, standardized some inconsistent or ambiguous values, and eliminated some outliers and errors in the data. These steps resulted in a clean and transformed dataset for the analysis. Table 1 shows a sample of the product features dataset after cleaning, along with their corresponding satisfaction scores.

Table 1: Sample of the cleaned product features dataset and satisfaction scores

| Product ($i$) | feature($j$) | | | | Overall score |
|---|---|---|---|---|---|
| | 1 | 2 | … | k | z |
| 1 | $f_{11}$ | $f_{12}$ | | $f_{1k}$ | $z_1$ |
| 2 | $f_{21}$ | $f_{22}$ | | $f_{2k}$ | $z_2$ |
| . | . | . | | . | . |
| . | . | . | | . | . |
| . | . | . | | . | . |
| m | $f_{m1}$ | $f_{m2}$ | | $f_{mk}$ | $z_m$ |

The product feature is indexed by j, where $j = 1,…,n$. The value of the j-th feature for the i-th product is denoted by $fij$, where $i = 1,…,m$. The overall satisfaction score for the i-th product is denoted by $z_i$.

## 3-2-Finding the best product features that affect the overall rating

A genetic algorithm (GA) is a method that mimics the process of natural evolution to find the best solution for a given problem. It works by creating and testing a group of possible solutions, called a population, and improving them over time based on their quality, or fitness (Chiong, 2009). This method can be used to solve different types of engineering problems, such as optimizing product design.
In this research, we use a GA to find the optimal combination of product features that affect the customer satisfaction score. The GA takes the product feature data as input and generates an initial population of candidate solutions, each consisting of a different subset of features. The GA then evaluates the fitness of each candidate solution by using the selected regression models, which are machine learning models that predict the customer satisfaction score.
Various machine learning algorithms, such as Decision Tree (DT), Random Forest (RF), K-Nearest Neighbors (KNN), Support Vector Regression (SVR), and Artificial Neural Network (ANN), are considered to find the best regression model as the fitness function of the GA. The fitness function evaluates the quality of each individual in the GA by using the regression model to predict the customer satisfaction score based on the product features. There are some studies that they developed different machine learing algorithms to measure and compare the performance of them in predicting customer satisfaction (Imtiaz & Ben Islam, 2020; Joung & Kim, 2023). The fitness function also uses k-fold cross validation, which is a technique that splits the data into k parts, trains the model on k-1 parts, and tests the model on the remaining part, to measure the performance of the regression model on unseen data (Géron, 2022). The fitness function repeats this process k times, using a different part as the test set each time, and calculates the average performance of the model on all the parts.

The fitness function uses various evaluation metrics to assess the performance of the model, such as, mean absolute error (MAE), mean squared error (MSE), and root mean squared error (RMSE), which are presented in Eqs. (1)–(3).

$$RMSE = \sqrt{\frac{\sum_{i=1}^{m}(z_i - \hat{z}_i)^2}{m}} \quad (1)$$

$$MAE = \frac{\sum_{i=1}^{m}|z_i - \hat{z}_i|}{m} \quad (2)$$

$$MSE = \frac{\sum_{i=1}^{m}(z_i - \hat{z}_i)^2}{m} \quad (3)$$

where $\hat{z}_i$ is predicted overall satisfaction score of a product, $z_i$ is actual overall satisfaction score and m is the number of products.

It is good to mention that the GA applies operators such as crossover, mutation, and selection to create new candidate solutions from the existing ones. These operators create new subsets of product features, as follows: First, the selection operator uses a roulette wheel selection, which is a method that assigns a probability of being selected to each candidate solution proportional to its fitness. The candidate solutions with higher fitness have higher chances of being selected as survivors. Then, on the survivors, the crossover operator and the mutation operator are applied. The crossover operator mixes the features of two parents, creating new feature combinations. The mutation operator alters the features of one mutant, creating new feature variations. The new candidate solutions are also evaluated by the regression models. The GA repeats this process until it finds the candidate solution that has the lowest prediction error (Chiong, 2009).

The final output of the GA is the best individuals, which represent the optimal combinations of product features. This way, we can offer product developers, marketers, and organizations different options of the best and most important subsets of features that they can work on to provide customer needs and satisfaction.

## 3-4-Determining importance values of product features on overall ratings

This study conducts SHAP (Lundberg and Lee, 2017), an interpretive machine learning technique, at the end of the selection process for the selected optimal subsets by genetic algorithm on the fitness function. Recent studies interpreted their ML models by SHAP (Deng et al., 2023; Joung & Kim, 2023; Park & Kim, 2024). SHAP analysis is a model-agnostic method that uses Shapley values, a concept from cooperative game theory, to quantify the contribution of each feature and each feature value to the model output, taking into account the interactions with other features.

The method proposed by Lundberg and Lee (2017) for calculating the SHAP values is shown in Eq. 4, where $\phi i$ is the SHAP value of feature $i$, which represents the average marginal effect of that feature on the prediction. $z'$ is a subset of input features $x'$, and M is the set of all features. $|z'|$ and $|M|$ are the cardinalities of $z'$ and $M$, respectively, meaning the number of elements in each set.

The equation compare the output value for subset $z'$ and the output when feature i is excluded from subset $z'$, i.e., $f_x(z') - f_x(z' \setminus i)$. The model evaluates the difference in the output made by feature $i$ for all combinations of features. To clarify, it basically sums up the differences between the predictions with and without feature $i$, weighted by the number of possible subsets of features, over all subsets of features. This way, the SHAP value captures the impact of adding a feature to the prediction, taking into account the interactions with other features.

$$\phi i = \sum_{z' \in x'} \frac{|z'|!(|M|-|z'|-1)!}{|M|!} [f_x(z') - f_x(z' \setminus i)] \tag{4}$$

As it mentioned before, we use a genetic algorithm to generate the best optimal subsets of features, that are the most effective combinations of product features for maximizing customer satisfaction. We then apply the SHAP analysis to the best selected individuals by GA to find the contribution of each feature.

The SHAP analysis uses a background dataset that serves as a reference for the feature values and simulates the effect of setting a particular feature to a "missing" or reference state. By replacing features with their background values, SHAP observes how changes in feature values impact the model's output. This simulation is crucial for understanding feature importance, especially when models may not handle missing data during testing.

SHAP returns a matrix of SHAP values, where each row corresponds to a product and each column corresponds to a feature. The SHAP values represent the contribution of each feature to the prediction for each product. Then, we calculate the average of the absolute sum of SHAP values for each feature and each specification of them across all products. The SHAP values are averaged over all rows to obtain the mean of SHAP value for each feature and each specification of the features. The mean SHAP value represents the effect of each feature and each specification of the features on the model output. The higher the mean SHAP value, the more positive the effect. The lower the mean SHAP value, the more negative the effect.

The mean SHAP values are used to understand customer preferences and satisfaction for product features and to drive product design and development. For example, the mean SHAP value for the feature "color" when it is "red" is the average of the SHAP values of all the instances that have "red" as their color. This indicates how much the color "red" influences the customer satisfaction score for the products. Algorithm 1 presents the steps followed to obtain importance of product features and different spec of them.

Algorithm 1

---

```
def GA(data, target, model):
t = 0
P = init_population()
evaluate(P, data, target, model)
while not done:
    t = t + 1
    P' = select_parents(P)
    recombine(P')
    mutate(P')
    evaluate(P', data, target, model)
    P = survive(P, P')
```

```
return P

def shap_value(P, data, model):
 shap_values = []
 shap_value = calculate_shap_value(data, model)
 shap_values.append(shap_value)
return shap_values
```

# 4-Implementation and results

This section illustrates the application of the proposed method to investigate the relative importance of different laptop attributes for customers. All implementation steps are performed using Spyder ver. 3.3.5 and Python ver. 3.7 software on a personal computer with the Linux OS, an Intel Core i9 processor (12th generation), and 128 GB of RAM. The data source is a Kaggle dataset that contains information about laptops sold on an e-commerce website (flipkart.com), where customers can post their comments and ratings on a 1–5 star scale(*Laptop data*, 2022). The dataset includes the product names, the product attributes, the overall ratings, and the number of reviews, which were collected using an automated chrome web extension tool called Instant Data Scraper.

The dataset consists of 579 products after filtering out those with less than 10 ratings. The product attributes was as follows: Brand (A1), Processor Brand Name (A2), Processor Name (A3), Processor Generation (A4), RAM Size (A5), RAM Type (A6), SSD Capacity (A7), HDD Capacity (A8), Operating System (A9), OS Bit (A10), Graphic Card Memory (A11), Weight (A12), Display Size (A13), Warranty (A14), Touchscreen (A15), MS Office (A16), Latest Price (A17), Old Price (A18), and Discount (A19),Reviews(A20), Ratings (a21), Star ratings(a22).

The categorical features are encoded using OrdinalEncoder, which is a method that converts the discrete values of the features into numerical values, such as 0, 1, 2, etc. The numeric features are scaled using MinMaxScaler, which is a method that transforms the numerical values of the features into a range between 0 and 1, so that they have the same scale and do not dominate the model.

## 4-1-Data collection and preprocessing

To prepare the data for the analysis, we removed some features that were either irrelevant or incomplete. The reviews feature was irrelevant because it did not provide any information about the laptop attributes, and the Generation of processor feature was incomplete because it had too many missing values. We also created a new feature, Mix rating, by combining the number of ratings and star rating features. The formula for the Mix rating feature is:

$$Mix\ rating = \frac{(\text{rating } \times \text{star rating})}{(\text{rating } + \ 1)}$$

According to the suggestion of statistical power, when inferring product qualities, people should pay attention not only to the average rating of the products but also to the number of people who rated them (Obrecht et al., 2007). Hoffart et al. (2019) Investigate how people use online ratings to make decisions about products or services. The authors propose a new model that incorporates both the average rating and the number of ratings as cues for quality and popularity. The model finds that people are sensitive to both cues. Therefore, we adopted this formula to account for both the quantity and quality factors of the ratings. This formula allocates more importance to the star rating, which indicates the overall satisfaction of customers with the product on a 1-5 scale, than to the rating, which represents the number of customers who gave the star ratings. The details of the collected data regarding laptops are shown in Table 2. This table includes some descriptive statistics of product attributes and mix ratings.

Table 2: descriptive statistics of product attributes and mix ratings

| Name | Mean | Mode | Median | Dispersion | Min | max |
|---|---|---|---|---|---|---|
| Brand | | ASUS | | 2.01 | | |
| Processor Name | | Core i5 | | 1.83 | | |
| RAM Size | | 8 | | 0.836 | | |
| RAM Type | | DDR4 | | 0.38 | | |
| SSD Capacity | | 512 | | 1.23 | | |
| HDD Capacity | | 0 | | 0.531 | | |
| Weight | | Casual | | 0.873 | | |
| Operating system | | Windows | | 0.167 | | |
| OS bit | | 64 | | 0.264 | | |
| Graphic card memory | | 0 | | 0.877 | | |
| Display size | | 15.6 | | 1.32 | | |
| Warranty | | 1 | | 0.776 | | |
| Touchscreen | | No | | 0.339 | | |
| MS Office | | No | | 0.68 | | |
| Discount | 19.58 | 23 | 20 | 0.53 | 0 | 57 |
| Old price | 82541.06 | 0 | 74990 | 0.50 | 0 | 326990 |
| Latest Price | 67242.8 | 54990 | 58994.5 | 0.52 | 19990 | 441990 |

| Mix ratings | 4.03503 | 4 | 4.18032 | 0.132298 | 1.275 | 4.79251 |

## 4-2- Determining the most influential product features and their significance on customer satisfaction scores

The aim of this study is to select and rank the most influential features for predicting customer satisfaction ratings of laptops sold on an e-commerce website. First, We applied a GA to find the optimal feature subset on the mentioned dataset with 17 features and one label, which is the mixed rating feature. Moreover, using feature selection algorithms can increase the accuracy of model (Chiong, 2009; Imtiaz & Ben Islam, 2020; Kuo et al., 2014; Mirjalili, 2019). the The data is preprocessed using pandas, sklearn and numpy libraries and encoded the categorical features. Furthermore, the GA parameters are set as follows: maximum number of iterations = 50, population size = 100, crossover rate = 0.9, mutation rate = 0.1. It is good to be noted that We favored the crossover operator over the mutation operator to exploit the best solutions and combine their features to create new solutions that are likely to be better or similar, while increasing the convergence speed and reducing the risk of harmful mutations (Chiong, 2009; Katoch et al., 2021).

In the second step, We initialized the population by randomly generating binary vectors of length 17, where each bit indicates the selection or exclusion of a feature. Each solution is evaluated with the fitness function of GA, which is a machine learning model that uses cross-validation, evaluation metrics.

Different machine learning models are developed as the fitness function for GA, such as RandomForestRegressor, DecisionTreeRegressor, and so on. Comparing their performance and results and finding out which one is better as the fitness function for GA. Based on the comparison, selecting the SVR model as the best fitness function for GA in this study. The SVR model uses the rbf kernel and the C and gamma parameters determined by continuous versions of GA. The fitness function performs 10-fold cross-validation, which is a method that splits the data into 10 parts, uses 9 parts for training and 1 part for testing, and repeats this process 10 times, each time using a different part for testing. The fitness function calculates MSE, MAE, and RMSE as the evaluation metrics, which are different ways of measuring the accuracy and performance of the predictions(Kelleher et al., 2020). It returns MAE as the fitness value for each solution, where the lower the MAE, the better the solution for the next generation. The performance of the fitness function for the proposed method is shown in Table 3.

Table 3: The performance of different fitness function for the proposed method

| The proposed method with different machine learning algorithms | MSE | RMSE | MAE |
|---|---|---|---|
| **KNN** | 0.271 | 0.521 | 0.321 |
| **SVR** | 0.225 | 0.475 | 0.297 |

| | | | | |
|---|---|---|---|---|
| **DT** | | 0.262 | 0.512 | 0.335 |
| **ANN** | | 0.272 | 0.521 | 0.348 |
| **RF** | | 0.235 | 0.485 | 0.325 |

The GA used roulette selection to choose solutions for the next generation based on their fitness values, while preserving some diversity in the population. It also applied crossover and mutation operations to generate new solutions(Mirjalili, 2019). It repeated this process for 50 generations and obtained the best feature subset and hyperparameters for each machine learning model. The output of the GA was a list of the best solutions and their evaluation metrics. We extracted the five best solutions, which are the five optimal feature subsets that affect the customer satisfaction ratings. We conducted SHAP analysis on fitness function for these five best solutions to rank the features in each subset and determine their importance.

We used the shap library to compute shaply values in our experiments. It is worth mentioning that the KernelExplainer class was used from the shap library, which can handle any machine learning model or python function. We take the average of the SHAP values across all products to get the mean SHAP value for each feature and each specification of the features.The mean SHAP value shows how each feature and each specification of the features affects the model output which is the customer satisfaction ratings. Next, we identified the most important feature for each solution and returned them as a list. Based on the results, SVR had the best performance. Therefore, The results in this study are for the proposed method with SVR as the fitness function.

The Beeswarm SHAP plots for the five optimal subsets of features, are shown in Figures 2 to 6. A beeswarm plot is a one-dimensional scatter plot that shows the distribution and importance of each feature for the model's output. The y-axis represents the features, and the x-axis represents the SHAP values, which measure the contribution of each feature to the prediction.The ranking of each feature in these optimal subsets are presented in Table 4. The shapv postfix in the figures is related to the calculated Shapley values corresponding to the features.The importance value for each feature and each spec of them for the first best solution based on shap values, is shown in Table 5.

Table 4: ranking of each feature in 5 selected optimal subsets of features

| Best | features | Ranking best features using shap model |
|---|---|---|
| 1 | 11 | A1 > A7 > A9 > A10 > A11 > A12 > A15 > A16 > A17 > A18 > A19 |
| 2 | 11 | A1 > A6 > A7 > A9 > A10 > A11 > A12> A16 > A17 > A18 > A19 |
| 3 | 10 | A1 > A7 > A10 > A11 > A12 > A15 > A16 > A17 > A18 > A19 |
| 4 | 12 | A1 > A6 > A7 > A9 > A10 > A11 > A12> A14 > A16 > A17 > A18 > A19 |
| 5 | 11 | A1 > A6 > A7 > A9 > A10 > A11 > A12 > A14 > A16 > A18 > A19 |

There are some studies that explore the importance of features and their different spec. However, they have limitations in finding the most important subsets of features or ranking them considering their interaction with each other when there is only user rating available. Our proposed method can fulfill all these issues by the result of table4. This result can help businesses to improve their product and marketing strategy based on customer preferences and to use their resources more efficiently by just working on important features. Moreover, it can identify common needs among customers which is a complicated task since the preferences of customers are different from one person to another person.

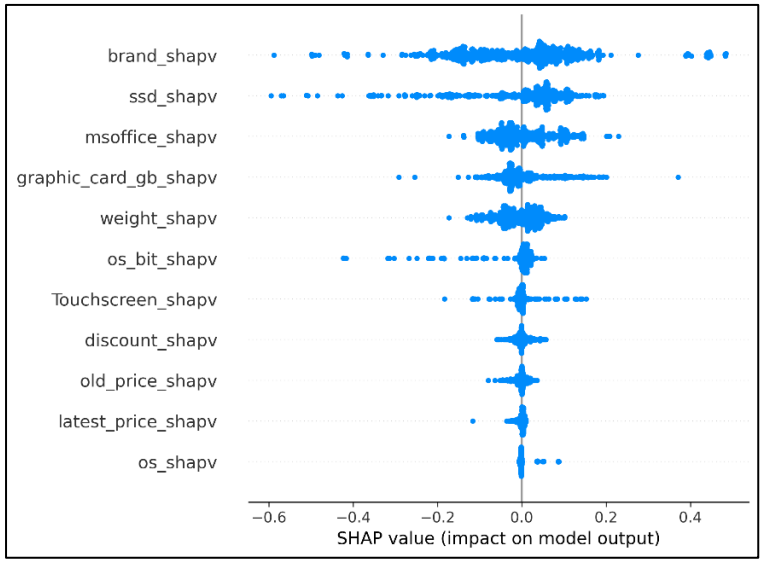

Figure 2: The Beeswarm SHAP plot for the 1st optimal subset

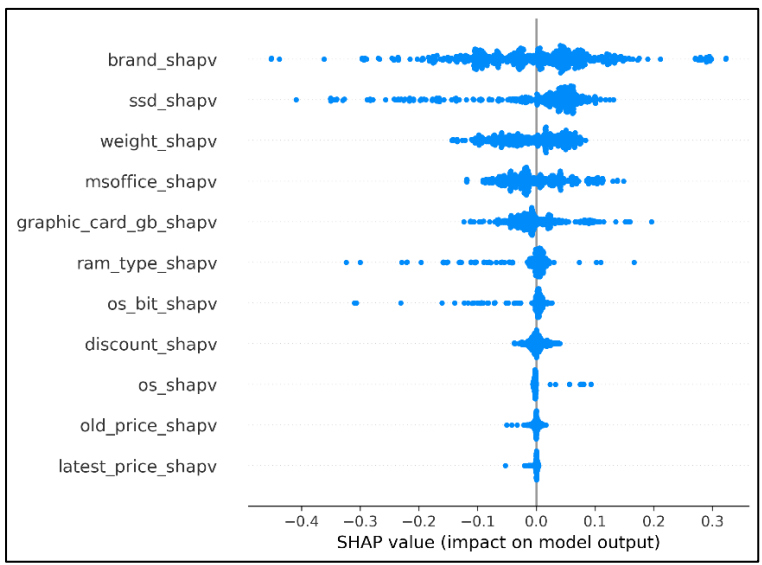

Figure 3: The Beeswarm SHAP plot for the 2nd optimal subset

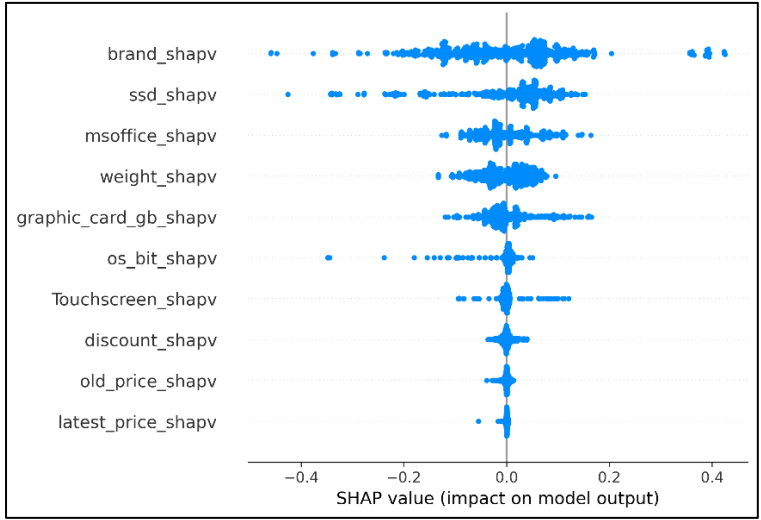

Figure 4: The Beeswarm SHAP plot for the 3rd optimal subset

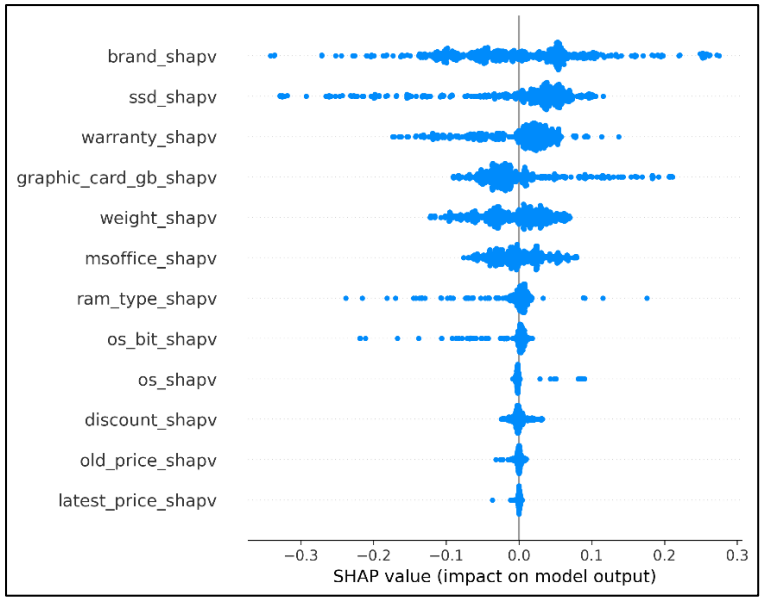

Figure 5: The Beeswarm SHAP plot for the 4th optimal subset

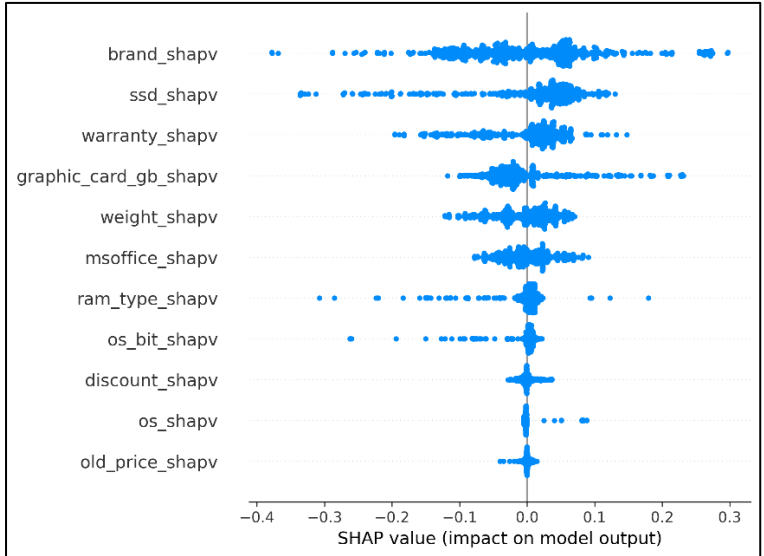

Figure 6: The Beeswarm SHAP plot for the 5th-optimal subset

Overall, brands in each optimal subsets have larger impacts on overall satisfaction. There are 5 optimal subsets that product developers can focus on each one of them that they like considering their values and resources.

Table5 shows a graphical representation of the SHAP result. The color bar on the right indicates feature values. The higher spec values are presented in high saturation, and the lower spec values are close to the white color with low saturation. The y-axis lists product features in the order of impact magnitude, and the x-axis represents the impact on the model output. In this study, the game theory ML approach predicts the customer's ratings among different products, so the graph explains the relationship between feature specs and customer satisfaction. Higher spec values have positive impacts on the customer's preferences. On the other hand, lower spec values have a negative effect on them.

Table 5: SHAP results for features' specs.

| brand | | Graphic card in gb | | Ssd | | weight | |
|---|---|---|---|---|---|---|---|
| acer | 0.007 | 0 gb | -0.017 | 0 | -0.235 | Casual | 0.018 |
| APPLE | 0.434 | 2 gb | -0.037 | 128 | 0.004 | Gaming | 0.022 |
| ASUS | 0.068 | 4 gb | 0.075 | 256 | 0.087 | ThinNlight | -0.037 |
| Avita | -0.337 | 6 gb | 0.109 | 512 | 0.037 | | |
| DELL | -0.121 | 8 gb | 0.036 | 1024 | -0.136 | | |
| HP | 0.045 | | | 2048 | -0.426 | | |
| Infinix | 0.084 | | | 3072 | -0.570 | | |

| | |
|---|---|
| Lenovo | -0.158 |
| LG | 0.055 |
| Mi | 0.079 |
| Microsoft | -0.051 |
| MSI | 0.106 |
| Nokia | 0.048 |
| Realme | 0.050 |
| RedmiBook | 0.100 |
| Vaio | -0.107 |

| msoffice | | os | | os_bit | | Touchscreen | |
|---|---|---|---|---|---|---|---|
| no | -0.038 | mac | 0.050 | 32 | -0.148 | no | -0.002 |
| yes | 0.049 | windows | -0.002 | 64 | 0.008 | yes | 0.013 |

| Latest Price | | Discount | | Old price | |
|---|---|---|---|---|---|
| 19990 | 0.007 | 0 | 0.012 | 26972 | 0.0014 |
| 21490 | 0.006 | 1 | 0.0003 | 27971 | 0.0135 |
| 23990 | 0.003 | 2 | 0.0353 | 30972 | 0.0012 |
| . | . | . | . | . | . |
| . | . | . | . | . | . |
| . | . | . | . | . | . |
| 226705 | -0.023 | 48 | 0.009 | 259715 | -0.0119 |
| 276704 | -0.031 | 56 | -0.054 | 273215 | -0.0019 |
| 441990 | -0.116 | 57 | -0.056 | 326990 | -0.0058 |

# 5-Validation and Disscusion

In this section, we present the comparisons of the rankings obtained by different methods and highlight the advantages of our proposed method. Support Vector Regression, K-Nearest Neighbors, Decision Tree, Random Forest, and Artificial Neural Network are created. We used them as standalone models without feature selection by GA. More detailed information on these model can be found in several studies(Géron, 2022; Kelleher et al., 2020; Sarker, 2021). Next, Ten fold cross validation is implemented for comparative study. This method splits data into ten equal width folds. Nine folds are trained to construct the model and the remaining fold is used to test it. This process is repeated ten times; thereby, the average performance metrics of the five models are taken. The performance metrics are mean squared error (MSE), root mean squared error (RMSE) and mean absolute error (MAE). The input for each model is the product attribute and the label is mix rating. The preprocessing for them is the same as the one for our proposed methodology. We also use Shap to determine the importance of features for each model. The only difference is that we do not use GA for these models. The results of the comparison are shown in Table 6. Based on the results, our proposed methodology has better performance than each one of these models in terms of mentioned performance metrics. It can also provide better, more and reliable design guidelines about product features.

Table 6: comparison of The proposed method with other standalone models

|  | MSE | RMSE | MAE |
|---|---|---|---|
| **The proposed method with using SVR as evaluate function** | 0.225 | 0.475 | 0.297 |
| **SVR** | 0.280 | 0.529 | 0.348 |
| **The proposed method with using KNN as evaluate function** | 0.271 | 0.521 | 0.321 |
| **KNN** | 0.306 | 0.553 | 0.366 |
| **The proposed method with using DT as evaluate function** | 0.262 | 0.512 | 0.335 |
| **DT** | 0.401 | 0.633 | 0.397 |
| **The proposed method with using ANN as evaluate function** | 0.272 | 0.521 | 0.348 |
| **ANN** | 0.371 | 0.609 | 0.412 |
| **The proposed method with using RF as evaluate function** | 0.235 | 0.485 | 0.325 |
| **RF** | 0.266 | 0.516 | 0.351 |

It is good to mention that among these created standalone models, SVR had the better result in terms of MAE and RF in terms of RMSE and MSE. According to the 10-fold cross-validation conducted in Python, the SVR model has an MAE of 0.348, RMSE of 0.529, and MSE of 0.280. The most important feature for the SVR model is brand (A1), followed by processor name (A3) and SSD (A7).The ranking of the features based on their SHAP values is A1> A3 > A7 > A13 > A16 > A12 > A11 > A14 > A5 > A2 > A8 > A10 > A6 > A15 > A19 > A9 > A18 > A17. The beeswarm SHAP plot for the SVR model is shown in Figure 7.

The proposed methodology is superior to the SVR model and the other standalone models in terms of  mentioned performance metrics and computational efficiency as well. A simulation can be conducted even if only the star ratings are available and the most important product features and their importance can be estimated. Therefore, we have performed a more comprehensive analysis with the proposed method.

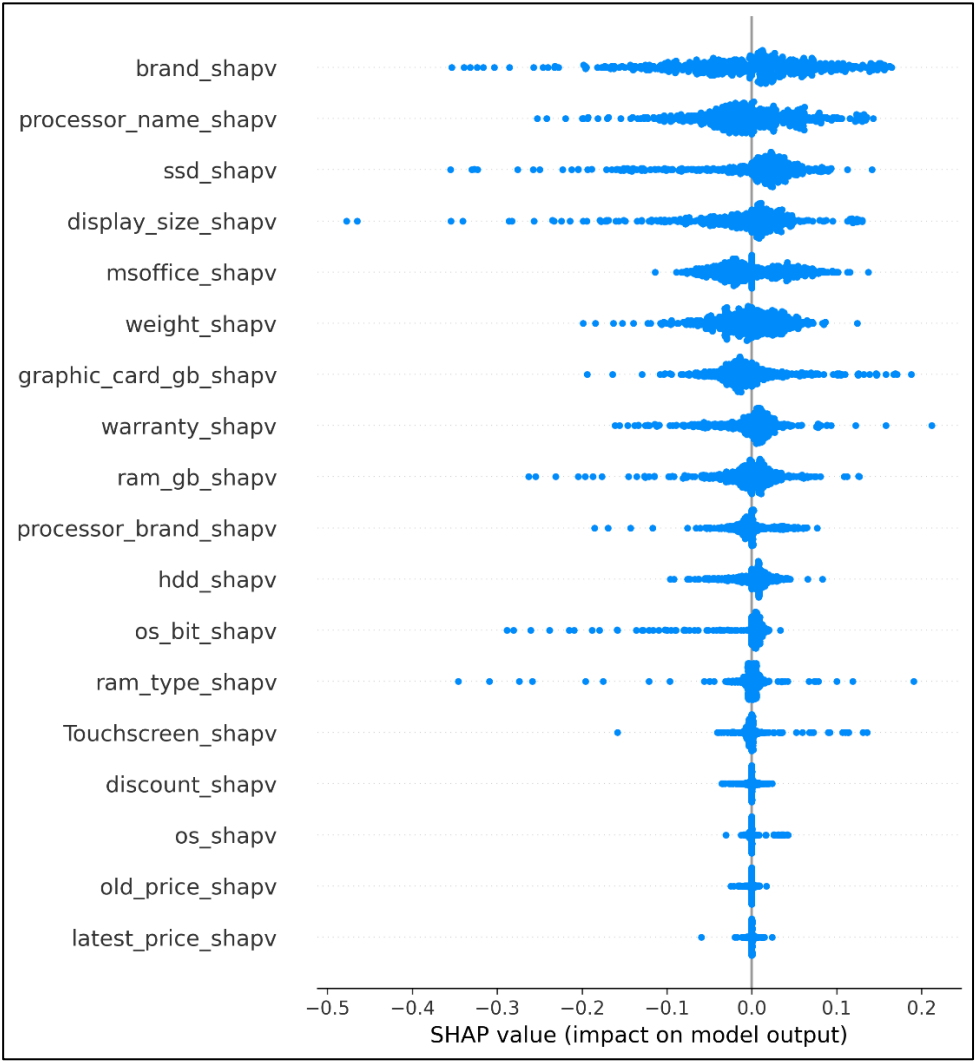

Figure 7: The beeswarm SHAP plot for the standalone SVR model

Moreover, Previous studies in this field have often ignored the joint impact of product features on customer satisfaction. In contrast, we use a genetic algorithm (GA) to explore the possibility of influencing customer satisfaction by all the features, even the least important ones. We also use a GA to select the best and most important subset of features that have the highest influence on customer satisfaction by considering their interrelationships. Some studies have only used online customer satisfaction data, such as reviews, to extract the impact of the features mentioned in this data and have overlooked the impact of other product features that may not have been mentioned. Moreover, some other studies have not ranked the features according to their importance and have only extracted the features. Furthermore, Some others have only obtained the importance of features and have not measured the degree of importance of each different spec of a feature. In this article, we use a GA to extract the best combinations of important and influential features on customer satisfaction and loyalty, which can be useful for product designers. They can use this information to choose the most important features and identify the best combination of these features. They can also decide which features to work on to improve their products or programs or services according to the organizational budget and resources. In addition, when they only have product feature data and satisfaction ratings, they can use our method to extract useful information and know the value of each feature and each different spec of them from the customer's point of view.

# 6-Conclusions

This research proposes a novel methodology based on artificial intelligence and online data to discover the importance of product attributes on customer satisfaction scores. The proposed method uses an explainable machine learning approach to extract the importance of product attributes from the overall ratings of products, and then uses a genetic algorithm to find the optimal combinations of product features that affect customer satisfaction. The method also uses the Shap model to interpret the model and provide design guidelines based on feature values and their interactions. The method is applied to a case study of laptops with data from 579 products. Based on our dataset, we acquire that brand has the highest impact on customer satisfaction. In addition, we conduct cross-validation between our proposed method, Support Vector Regression, K-Nearest Neighbors, Decision Tree, Random Forest, and Artificial Neural Network. As a result, we get better performance scores by employing the proposed game theory ML model. There are several studies that explore online user data for developing product features. However, the resultant implications of these studies are suitable for feature extraction, but not for feature ranking, selection, or interaction. Besides, companies need strategies to deal with product ratings only, and to determine the importance of features and their best combination. This study bridged this gap by proposing an explainable machine learning approach to extract comprehensive design strategies from online user-generated data. Specifically, the proposed methodology provide these informations:

- determining best optimal subsets of features, considering their interaction with each other
- Trains an explainable machine learning model that predicts customer ratings and interprets the model using SHAP.
- Draws comprehensive design implications by analyzing the SHAP result.

The result provided spec guidance for the important features. In summary, this research demonstrates the value and applicability of data-driven product design using artificial intelligence and online data. The method can also provide product designers and marketers with useful information and guidelines to optimize products and market them effectively.

## 6-1-Limitations and future research

The proposed method has some limitations that can be addressed in future research. First, the method can be applied to other domains and products, such as cameras, mobile phones, etc., to test its generalizability and robustness. Second, the method can use interactive evolutionary algorithms to involve humans in the optimization process and find better and more desirable solutions.

## 6-2-Practical implications and theoretical contributions

The results of this research can fill the research gap in the fields of customer satisfaction, online data analysis, multi-criteria decision making, and data-driven product design. The article by Sadaf Chali states that there is a research gap in determining the weights of product features according to user preferences in situations where only the overall customer satisfaction scores are available (Sadaf Chali 2023). this gap in the field of multi-criteria decision making (MCDM) is addressed by the method of our study.
The proposed method can help companies to produce suitable products that meet or exceed customer expectations, by identifying the most important and valuable features of a product and their importance. The method can also help companies to design and implement more effective marketing strategies, such as optimizing pricing, promotion of product features and so on. therefore, it can help to strengthen relationships with customers and increase customer retention and loyalty.The study also explains the nature of the black-box machine learning model and its predictions in a transparent and interpretable way. In other words, this study demonstrates the importance of an explainable ML model for data-driven information systems (IS) and information management (IM) research, as it helps to interpret the nonlinear relationships between independent and dependent variables in various prediction models. The study also shows the potential of the proposed method for new product development and marketing applications, as it reveals the influence of product features on customer satisfaction.

Abkenar, S. B., Kashani, M. H., Mahdipour, E., & Jameii, S. M. (2021). Big data analytics meets social media: A systematic review of techniques, open issues, and future directions. *Telematics and informatics*, *57*, 101517. https://doi.org/https://doi.org/10.1016/j.tele.2020.101517

Angelopoulos, S., Brown, M., McAuley, D., Merali, Y., Mortier, R., & Price, D. (2021). Stewardship of personal data on social networking sites. *International Journal of Information Management*, *56*, 102208. https://doi.org/https://doi.org/10.1016/j.ijinfomgt.2020.102208

Bashir, N., Papamichail, K. N., & Malik, K. (2017). Use of social media applications for supporting new product development processes in multinational corporations. *Technological Forecasting and Social Change*, *120*, 176-183. https://doi.org/https://doi.org/10.1016/j.techfore.2017.02.028

Bhimani, H., Mention, A.-L., & Barlatier, P.-J. (2019). Social media and innovation: A systematic literature review and future research directions. *Technological Forecasting and Social Change*, *144*, 251-269. https://doi.org/https://doi.org/10.1016/j.techfore.2018.10.007

Bi, J.-W., Liu, Y., Fan, Z.-P., & Cambria, E. (2019). Modelling customer satisfaction from online reviews using ensemble neural network and effect-based Kano model. *International Journal of Production Research*, *57*(22), 7068-7088. https://doi.org/https://doi.org/10.1080/00207543.2019.1574989

Bi, J.-W., Liu, Y., Fan, Z.-P., & Zhang, J. (2019). Wisdom of crowds: Conducting importance-performance analysis (IPA) through online reviews. *Tourism management*, *70*, 460-478. https://doi.org/https://doi.org/10.1016/j.tourman.2018.09.010

Briard, T., Jean, C., Aoussat, A., & Véron, P. (2023). Challenges for data-driven design in early physical product design: A scientific and industrial perspective. *Computers in Industry*, *145*, 103814. https://doi.org/https://doi.org/10.1016/j.compind.2022.103814

Çalı, S., & Baykasoğlu, A. (2022). A Bayesian based approach for analyzing customer's online sales data to identify weights of product attributes. *Expert Systems with Applications*, *210*, 118440. https://doi.org/https://doi.org/10.1016/j.eswa.2022.118440

Chiong, R. (2009). *Nature-inspired algorithms for optimisation* (Vol. 193). Springer.

Dash, G., Kiefer, K., & Paul, J. (2021). Marketing-to-Millennials: Marketing 4.0, customer satisfaction and purchase intention. *Journal of business research*, *122*, 608-620. https://doi.org/https://doi.org/10.1016/j.jbusres.2020.10.016

de Camargo Fiorini, P., Seles, B. M. R. P., Jabbour, C. J. C., Mariano, E. B., & de Sousa Jabbour, A. B. L. (2018). Management theory and big data literature: From a review to a research agenda. *International Journal of Information Management*, *43*, 112-129. https://doi.org/https://doi.org/10.1016/j.ijinfomgt.2018.07.005

Deng, S., Zhu, Y., Duan, S., Yu, Y., Fu, Z., Liu, J., Yang, X., & Liu, Z. (2023). High-frequency forecasting of the crude oil futures price with multiple timeframe predictions fusion. *Expert Systems with Applications*, *217*, 119580. https://doi.org/https://doi.org/10.1016/j.eswa.2023.119580

Deng, Z., Lu, Y., Wei, K. K., & Zhang, J. (2010). Understanding customer satisfaction and loyalty: An empirical study of mobile instant messages in China. *International Journal of Information Management*, *30*(4), 289-300. https://doi.org/https://doi.org/10.1016/j.ijinfomgt.2009.10.001

Du, Y., Liu, D., Morente-Molinera, J. A., & Herrera-Viedma, E. (2022). A data-driven method for user satisfaction evaluation of smart and connected products. *Expert Systems with Applications*, *210*, 118392. https://doi.org/https://doi.org/10.1016/j.eswa.2022.118392

Géron, A. (2022). *Hands-on machine learning with Scikit-Learn, Keras, and TensorFlow*. " O'Reilly Media, Inc.".

Groves, R. M. (2006). Nonresponse rates and nonresponse bias in household surveys. *International Journal of Public Opinion Quarterly*, *70*(5), 646-675. https://doi.org/https://doi.org/10.1093/poq/nfl033

Guo, Y., Barnes, S. J., & Jia, Q. (2017). Mining meaning from online ratings and reviews: Tourist satisfaction analysis using latent dirichlet allocation. *Tourism management*, *59*, 467-483. https://doi.org/https://doi.org/10.1016/j.tourman.2016.09.009

Haleem, A., Javaid, M., Qadri, M. A., Singh, R. P., & Suman, R. (2022). Artificial intelligence (AI) applications for marketing: A literature-based study. *International Journal of Intelligent Networks*. https://doi.org/https://doi.org/10.1016/j.ijin.2022.08.005

Heidemann, J., Klier, M., & Probst, F. (2012). Online social networks: A survey of a global phenomenon. *Computer networks*, *56*(18), 3866-3878. https://doi.org/https://doi.org/10.1016/j.comnet.2012.08.009

Hoffart, J. C., Olschewski, S., & Rieskamp, J. (2019). Reaching for the star ratings: A Bayesian-inspired account of how people use consumer ratings. *Journal of Economic Psychology*, *72*, 99-116. https://doi.org/https://doi.org/10.1016/j.joep.2019.02.008

Hou, T., Yannou, B., Leroy, Y., & Poirson, E. (2019). Mining customer product reviews for product development: A summarization process. *Expert Systems with Applications*, *132*, 141-150. https://doi.org/https://doi.org/10.1016/j.eswa.2019.04.069

Imtiaz, M. N., & Ben Islam, M. K. (2020). Identifying significance of product features on customer satisfaction recognizing public sentiment polarity: Analysis of smart phone industry using machine-learning approaches. *Applied Artificial Intelligence*, *34*(11), 832-848. https://doi.org/https://doi.org/10.1080/08839514.2020.1787676

Jeong, B., Yoon, J., & Lee, J.-M. (2019). Social media mining for product planning: A product opportunity mining approach based on topic modeling and sentiment analysis. *International Journal of Information Management*, *48*, 280-290. https://doi.org/https://doi.org/10.1016/j.ijinfomgt.2017.09.009

Joung, J., & Kim, H. (2023). Interpretable machine learning-based approach for customer segmentation for new product development from online product reviews. *International Journal of Information Management*, *70*, 102641. https://doi.org/https://doi.org/10.1016/j.ijinfomgt.2023.102641

Joung, J., Kim, K.-H., & Kim, K. (2021). Data-driven approach to dual service failure monitoring from negative online reviews: managerial perspective. *SAGE Open*, *11*(1), 2158244020988249. https://doi.org/https://doi.org/10.1177/2158244020988249

Kalro, H., & Joshipura, M. (2023). Product attributes and benefits: integrated framework and research agenda. *Marketing Intelligence & Planning*, *41*(4), 409-426. https://doi.org/https://doi.org/10.1108/MIP-09-2022-0402

Kano, N. (1984). Attractive quality and must-be quality. *Journal of the Japanese society for quality control*, *31*(4), 147-156.

Katoch, S., Chauhan, S. S., & Kumar, V. (2021). A review on genetic algorithm: past, present, and future. *Multimedia tools and applications*, *80*, 8091-8126. https://doi.org/https://doi.org/10.1007/s11042-020-10139-6

Kelleher, J. D., Mac Namee, B., & D'arcy, A. (2020). *Fundamentals of machine learning for predictive data analytics: algorithms, worked examples, and case studies*. MIT press.

Kim, H.-S., & Noh, Y. (2019). Elicitation of design factors through big data analysis of online customer reviews for washing machines. *Journal of Mechanical Science and Technology*, *33*, 2785-2795. https://doi.org/https://doi.org/10.1007/s12206-019-0525-5

Kim, H. H. M., Liu, Y., Wang, C. C., & Wang, Y. (2017). Data-driven design (D3). *Journal of Mechanical Design*, *139*(11), 110301. https://doi.org/https://doi.org/10.1115/1.4037943

Kuo, R., Chiang, N., & Chen, Z.-Y. (2014). Integration of artificial immune system and K-means algorithm for customer clustering. *Applied Artificial Intelligence*, *28*(6), 577-596. https://doi.org/https://doi.org/10.1080/08839514.2014.923167

*Laptop data*. (2022). https://www.kaggle.com/datasets/kuchhbhi/2022-march-laptop-data

Li, S., Zhang, Y., Li, Y., & Yu, Z. (2021). The user preference identification for product improvement based on online comment patch. *Electronic Commerce Research*, *21*, 423-444. https://doi.org/https://doi.org/10.1007/s10660-019-09372-5

Li, Y., Dong, Y., Wang, Y., & Zhang, N. (2023). Product design opportunity identification through mining the critical minority of customer online reviews. *Electronic Commerce Research*, 1-29. https://doi.org/https://doi.org/10.1007/s10660-023-09683-8

Mirjalili, S. (2019). Evolutionary algorithms and neural networks. In *Studies in computational intelligence* (Vol. 780). Springer.

Nilashi, M., Abumalloh, R. A., Samad, S., Alrizq, M., Alyami, S., & Alghamdi, A. (2023). Analysis of customers' satisfaction with baby products: The moderating role of brand image. *Journal of Retailing and Consumer Services*, *73*, 103334. https://doi.org/https://doi.org/10.1016/j.jretconser.2023.103334

Noori, B. (2021). Classification of customer reviews using machine learning algorithms. *Applied Artificial Intelligence*, *35*(8), 567-588. https://doi.org/https://doi.org/10.1080/08839514.2021.1922843

Obrecht, N. A., Chapman, G. B., & Gelman, R. (2007). Intuitive t tests: Lay use of statistical information. *Psychonomic bulletin & review*, *14*(6), 1147-1152. https://doi.org/https://doi.org/10.3758/BF03193104

Park, S., & Kim, H. (2024). Extracting product design guidance from online reviews: An explainable neural network-based approach. *Expert Systems with Applications*, *236*, 121357. https://doi.org/https://doi.org/10.1016/j.eswa.2023.121357

Purnama, D. A., & Masruroh, N. A. (2023). Online data-driven concurrent product-process-supply chain design in the early stage of new product development. *Journal of Open Innovation: Technology, Market, and Complexity*, *9*(3), 100093. https://doi.org/https://doi.org/10.1016/j.joitmc.2023.100093

Roelen-Blasberg, T., Habel, J., & Klarmann, M. (2023). Automated inference of product attributes and their importance from user-generated content: Can we replace traditional market research? *International Journal of Research in Marketing*, *40*(1), 164-188. https://doi.org/https://doi.org/10.1016/j.ijresmar.2022.04.004

Sarker, I. H. (2021). Machine learning: Algorithms, real-world applications and research directions. *SN computer science*, *2*(3), 160. https://doi.org/https://doi.org/10.1007/s42979-021-00592-x

Suryadi, D., & Kim, H. M. (2019). A data-driven methodology to construct customer choice sets using online data and customer reviews. *Journal of Mechanical Design*, *141*(11), 111103. https://doi.org/https://doi.org/10.1115/1.4044198

Wang, J., Lai, J.-Y., & Lin, Y.-H. (2023). Social media analytics for mining customer complaints to explore product opportunities. *Computers & Industrial Engineering*, *178*, 109104. https://doi.org/https://doi.org/10.1016/j.cie.2023.109104

Wang, Y., Lu, X., & Tan, Y. (2018). Impact of product attributes on customer satisfaction: An analysis of online reviews for washing machines. *Electronic Commerce Research and Applications*, *29*, 1-11. https://doi.org/https://doi.org/10.1016/j.elerap.2018.03.003

Yang, B., Liu, Y., Liang, Y., & Tang, M. (2019). Exploiting user experience from online customer reviews for product design. *International Journal of Information Management*, *46*, 173-186. https://doi.org/https://doi.org/10.1016/j.ijinfomgt.2018.12.006

Yang, T., Dang, Y., & Wu, J. (2023). How to prioritize perceived quality attributes from consumers' perspective? Analysis through social media data. *Electronic Commerce Research*, 1-29. https://doi.org/https://doi.org/10.1007/s10660-022-09652-7

Zhou, L., Tang, L., & Zhang, Z. (2023). Extracting and ranking product features in consumer reviews based on evidence theory. *Journal of Ambient Intelligence and Humanized Computing*, *14*(8), 9973-9983. https://doi.org/https://doi.org/10.1007/s12652-021-03664-1